# Cross-Model Image Annotation Platform with Active Learning

Ng Hui Xian Lynnette[1], Henry Ng Siong Hock[1,2], and Nguwi Yok Yen[2]

[1] Defence Science Technology Agency, Singapore,
`{nhuixia1, hngsiong}@dsta.gov.sg`
[2] National Technological University, Singapore, `yokyen@ntu.edu.sg`

**Abstract.** We have seen significant leapfrog advancement in machine learning in recent decades. The central idea of machine learnability lies on constructing learning algorithms that learn from good data. The availability of more data being made publicly available also accelerates the growth of AI in recent years. In the domain of computer vision, the quality of image data arises from the accuracy of image annotation. Labelling large volume of image data is a daunting and tedious task. This work presents an End-to-End pipeline tool for object annotation and recognition aims at enabling quick image labelling. We have developed a modular image annotation platform which seamlessly incorporates assisted image annotation (annotation assistance), active learning and model training and evaluation. Our approach provides a number of advantages over current image annotation tools. Firstly, the annotation assistance utilizes reference hierarchy and reference images to locate the objects in the images, thus reducing the need for annotating the whole object. Secondly, images can be annotated using polygon points allowing for objects of any shape to be annotated. Thirdly, it is also interoperable across several image models, and the tool provides an interface for object model training and evaluation across a series of pre-trained models. We have tested the model and embeds several benchmarking deep learning models. The highest accuracy achieved is 74%.

## 1 Introduction

Artificial Intelligence (AI) was first coined by John McCarthy in 1956 ([5]). Arthur Samuel then popularized the term Machine Learning (ML) in 1959 ([9]). The late 1950s was a period where AI founders started to ponder the possibility of machine to take over the human decision-making process by mimicking the way human thinks and learns. Artificial intelligence revolves around making machines exhibit human like decision making processes, while machine learning enables computers to learn on its own without explicit programming.

Despite these good starting points of AI, it is only until recent decade that we see a significant leapfrog advancement in this field. The central idea of machine learning is the machine's ability to learn from data. The availability of larger volumes of publicly available data together with open-source models accelerates the growth of the field of AI, particularly in training the machine to learn through supervised learning.

The success of object classification and recognition models drive researchers' interest in applying these models in a wide range of new application domains. In order to harness the appropriate model and hardware for the domain, we need to quickly bootstrap the annotated corpora and perform model training and evaluation. This can only be done with the availability of labelled data as ground truth, with annotation done by domain expert. The labelling process does not come easy, especially in the field of computer vision where the machine is trained to recognize objects within an image. This calls for object to be recognized, categorized and mapped to the appropriate word(s) to describe the object meaningfully. Thus, we identify three requirements image annotation tool should have in order to meet today's data labelling demands:

**Annotation assistance.** Creating domain-specific annotated dataset is extremely challenging and requires experts who are familiar with the objects in the images in order to have good quality annotations. These domains typically have specialized image datasets that are not captured in the ImageNet ([1]) or CoCo dataset ([4]). To improve the quality of the tasks, annotation tool should be able to host reference images so that human annotator can identify similar objects in the image. It should also have a quality check module for annotators to edit the drawn annotated points or the labels, in order to maintain a high inter-annotator agreement.

**Interoperability between models.** There are many open source object classification and recognition models available for researchers to train their specialized datasets. The annotation tool needs to generate annotation points that can be used across a series of models. Each model takes in annotation data and output to different formats, and the annotation tool should be able to perform the additional pre-processing steps required to format the annotation output to the required for model training.

**Model training and evaluation.** The annotation tool should be able to perform model training on a suite of models and evaluate their accuracy. Different model platforms provide different types of visualization tools for model evaluation – Tensorflow uses Tensorboard, pyTorch uses Visdom – and it would be useful for a consolidated board to free up researchers from manually examining comparisons across platforms. Being able to evaluate different models from edge models like MobileNet ([3]) to larger model like ResNet101 ([14]) allows researchers to understand the performance of each model to suit their future deployment limitations like hardware limits or GPU memory limits. Knowing the accuracy of each model relative to the other models allows deployment teams to make better decisions on deployment set up and model accuracy.

We developed an end-to-end annotation pipeline tool to provide a web-based interface for user to focus on labelling data and evaluating results from several models. Reference knowledge base of images and text can be input in the system for the user to refer during annotation process, assisting in their identification of objects. An active learning feature reduces the need for manually labelling each image. The labelled data format is standardized and the necessary pre-processing scripts are integrated to allow the images to be used with popular image classification algorithms like Tensorflow ([2]) and Pytorch ([8]). The modular architecture of this work enables users to augment their instance with custom datasets, machine learning algorithms, reference knowledge base and annotation types. In the deployment of this work, users annotate on aver- age 52 images per hour, as compared to 31 images per hour with other image annotation tools.

## 2  Related Work

In recent years, several image annotation tools have been developed. Some are open-sourced while some are paid tools, but none offer an integrated environment for annotation and multiple model evaluation.

Prodigy ([7]) provides a web-based interface to annotate visual objects in four different ways: (1) draw a boundary and add a tag to objects in images, (2) accept or reject a labelled object, (3) select multiple labels to an image, and (4) select multiple images for a given label. Researchers can write their own Prodigy recipes to configure the labels and image directories and integrate with Tensorflow's object detection API. LabelMe ([13]) is an image polygonal annotation tool built with Python that can be used for image segmentation. Both Prodigy and LabelMe do not support annotation assistance, integrated data export formats and model training out of the box.

The general approach taken by Microsoft's Visual Object Tagging Tool ([6]) and Intel's CVAT ([10]) is very similar to our approach. However, they focus strongly on a single-annotator use case rather than allowing concurrent annotators to work on a single file directory. While VOTT and CVAT can export annotated data in a variety of formats to fit different models, it decouples model training and evaluation.

## 3  Annotation Tool

This work offers a number of functionalities expected from a generic image annotation platform: an intuitive user interface, flexible configuration of the annotation classes and the ability to run multiple instances concurrently for multiple users.

In this paper, we focus on the this work unique features: (1) annotation assistance via reference hierarchy and images, quality check and active learning; (2) interoperability between image models via modular design for encompassing different models; and (3) model training and evaluation to understand model performance on different subsets of the data.

## 3.1 System Overview

This work is implemented with a ReactJS [3] web application front end and a Python Flask [4] as a backend server host. ReactJS facilitates building encapsulated components which is essential for a modular architecture. For backend modules, Python is the language of choice as many machine learning algorithms have Python-based implementation. Flask is chosen as the host server for its tight integration with Python and its facilitation of a server-based applications with its ability to scale.

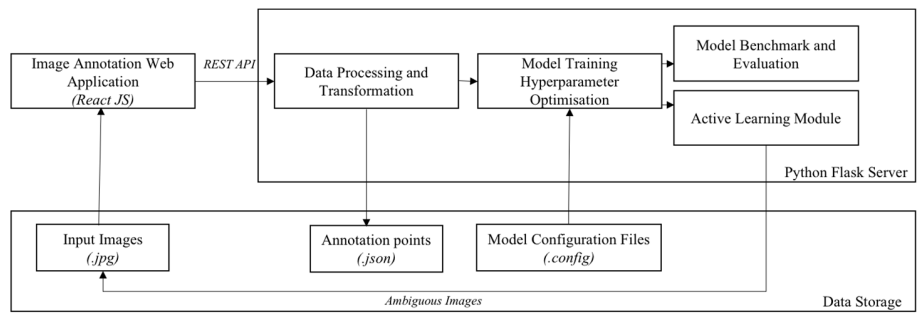

Fig. 1: System Overview

## 3.2 Annotation User Interface

This work presents a simple to access web browser-based interface, with most of its code hosted on a server. This allows multiple users to access the system at the same time. If multiple users were annotating the same folder, they will be presented with different images, because an image lock system is implemented. Should there be no movement on the same image after 30 minutes, the image is released for annotation by another user. To reduce annotation efforts, users can annotate multiple objects in a single image.

Figure 2 presents a flow of the User Interface for this work annotation tool. The annotation interface opens with (1), a list of folders where there are images to be annotated. Upon selecting a folder, screen (2) appears, where the user annotates the images by drawing polygons around the desired objects. To assist the user in selecting the appropriate label for the image, a (3) reference hierarchy drop down menu is implemented to narrow the image class, and (4) reference images are displayed where available. These reference images can be linked via knowledge databases or folder linking. To ensure high quality annotations, (5) a quality check module is implemented for users to reject the annotations if required.

---
[3] ReactJS: https://reactjs.org  
[4] Python Flask: https://github.com/pallets/flask

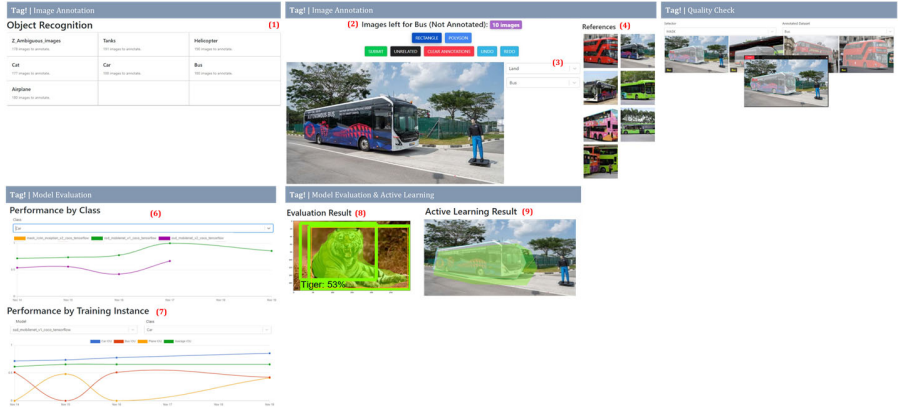

Fig. 2: Annotation Interface

In model evaluation, (6) shows the performance per model for each training instance over time while (7) shows the performance of each class over time on a certain model. This evaluates whether some models are better in detecting certain types of objects, and if training performance has plateaued out, indicating that further annotation and training efforts will have minimal impact on model performance. Part (8) depicts a typical training result on an object, where the green box and mask is the output from the model, along with the indicated class and percentage of confidence. The green model output is contrasted with the orange human annotation. (9) Active learning is implemented that shows the objects that the model is highly confident of, and do not require human annotation.

### 3.3 Annotations Assistance

This work reduces the cognitive load of annotators by assisting the annotation process, enabling users to annotate faster.

**Reference hierarchy and images.** Most of the images that require annotations are uncommon images requiring specialised domain knowledge. However, it is very difficult to find annotators with the domain knowledge, hence we built a module to allow researchers to link the annotation tool to a reference image directory. Annotators can then compare between the reference image and the actual image to determine the image class. A reference hierarchy is implemented as drop down modules for users to select the class of object in a more methodological fashion, subdividing categories of images further for user to find the desired class more easily.

**Quality check.** After annotators submit their annotations, it will be presented at the Quality Check page. Annotators can delete or edit annotation points, or edit annotation labels in the case of mistakes in labelling, thus ensuring the models are trained on high quality annotations.

**Active learning.** Active learning is a semi-supervised machine learning approach, in which the goal is to allow the machine to choose which data it wants to learn from to better improve its accuracy. We use the properties of active learning to prioritise images for humans to annotate. The system uses an uncertainty sampling strategy ([11]) working on the final image classification layer that produces a confidence score of classifying an object in an image to the corresponding label. When the model is 80% confident of the object's label, the annotation produced by the model will be added to the annotated corpora, with provision for human annotators to reject the annotation at the Quality Check page. Images that are within 40-60% of the label classification will be prioritised for human annotators to identify. Over time, as more images are annotated, the models are better trained, and will reduce the cognitive workload on human annotators by sieving out ambiguous images to annotate.

### 3.4 Interoperability of image models

After annotation step, a data augmentation step is performed. It generates variants and angles of a single image. Image processing techniques such as random flip, crops or resizing are used to modify the images through the modification of the shape of the image pixel array. This is crucial as most specialised objects do not have many photographs available.

In training the models on the specialised annotated datasets, transfer learning is usually used to leverage on the pre-trained models that have trained on large image datasets like ImageNet ([1]), which includes 1000 image classes.

This work has a modular support to include additional pre-trained models from open-source code, and perform the necessary data transformation required to fit the annotation data into the model's format. Further training parameters like learning rates and number of epochs can be configured in model configuration files. Several configuration files can be written for different optimisation parameters. Researchers may write their own data transformation scripts should there be a need to support a model that requires a different format. Currently, the pre-trained models supported are: SSD MobileNet V1 COCO Tensorflow, SSD MobileNet V2 COCO Tensorflow, Mask RCNN Inception V2 COCO Tensorflow, Mask RCNN Resnet50 Atrous COCO Tensorflow. These models are downloaded from the Tensorflow Model Zoo[12].

## 4 Model Training and Evaluation

We have developed four models. A series of images are annotated by drawing polygon points around the object of interest, and all the models are ran through it.

Researchers can select the entire dataset for training and evaluation or subsets of the data to build specific models. The annotated dataset that is selected for training is then split into 80% for training and 20% for evaluation. The generated evaluation result is then validated against the annotated images that are labelled by human annotators. The metric used in evaluating the effectiveness of the model is Intersection over Union (IoU), which measures the area of overlap between the mask produced by the model and the polygon annotated by the human annotator.

Table 1 presents the model evaluation results, showing how different models perform to different classes of data. The different performance of the models may be due to inherent properties of the images and objects, like background images or aeroplane wings. The results of our model cannot be benchmarked at the moment as the images being used in this work are not found in the standard or COCO dataset.

Table 1: Evaluation of Model Results. We note that the model results are not equivalent to the SOTA results as the images trained upon are not found in the COCO dataset, and hence are of very little quantity.

| Object Name | SSD MobileNet V1 COCO Tensorflow | SSD MobileNet V2 COCO Tensorflow | Mask RCNN Inception V2 COCO Tensorflow | Incep-Resnet50 COCO Tensorflow |
|---|---|---|---|---|
| Airborne vehicles | 0.42 | 0.43 | 0.54 | 0.64 |
| Ground vehicles | 0.65 | 0.72 | 0.74 | 0.74 |
| Rotorcrafts | 0.25 | 0.50 | 0.62 | 0.72 |

## 5   Conclusion and Future Work

We have presented this work as the first modular image annotation platform which seamlessly incorporates image annotation with annotation assistance, active learning and model training and evaluation. Our approach provides a number of advantages over current image annotation tools. Annotation assistance via a reference hierarchy and reference images greatly aid annotators in identifying the objects in the images, especially for domain-specific objects; thus reducing the need for specialized annotators to annotate the objects, but opens the possibility for the annotations to be done by anyone. Active learning prioritizes the images to annotate, reducing the cognitive load on the annotators. The modular approach used in this work allows the incorporation of new image classification algorithms, and the model training and evaluation pipeline provides statistics on the performance of the different algorithms across different subsets of the data. This work will be continually improved to align the platform to the needs of the community of researchers and annotators. Image data is being collected and annotated, to use in further statistical validation. Some areas that are currently being worked on: incorporation of new image annotation models and training statistics to allow for better evaluation of the models; improving the active learning through different sampling methods; adding support for multiple annotations on the same image and calculating inter-annotator support. Further investigation will be done to understand how image properties affect the model results.